\documentclass{ecai} 

\usepackage{algorithm}
\usepackage{algpseudocode}
\usepackage{makecell}
\usepackage{xcolor}
\usepackage{latexsym}
\usepackage{amsmath,amsfonts,bm,amssymb}
\usepackage{amsthm}
\usepackage{booktabs}
\usepackage{enumitem}
\usepackage{graphicx}
\usepackage{color}
\usepackage{bibentry}
\usepackage{xspace}
\newcommand{\hybrid}{combinatorial and mixed\xspace}

\newcommand{\BibTeX}{B\kern-.05em{\sc i\kern-.025em b}\kern-.08em\TeX}

\begin{document}


\begin{frontmatter}


\paperid{4477} 


\title{MOCA-HESP: Meta High-dimensional Bayesian Optimization for Combinatorial and Mixed Spaces via Hyper-ellipsoid Partitioning}


\author[A]{\fnms{Lam}~\snm{Ngo}\thanks{Corresponding Author. Email: s3962378@student.rmit.edu.au}}
\author[A]{\fnms{Huong}~\snm{Ha}}
\author[A]{\fnms{Jeffrey}~\snm{Chan}} 
\author[B]{\fnms{Hongyu}~\snm{Zhang}} 

\address[A]{RMIT University, Australia}
\address[B]{Chongqing University, China}


\begin{abstract}
High-dimensional Bayesian Optimization (BO) has attracted significant attention in recent research. However, existing methods have mainly focused on optimizing in continuous domains, while combinatorial (ordinal and categorical) and mixed domains still remain challenging. In this paper, we first propose MOCA-HESP, a novel high-dimensional BO method for combinatorial and mixed variables. The key idea is to leverage the hyper-ellipsoid space partitioning (HESP) technique with different categorical encoders to work with high-dimensional, combinatorial and mixed spaces, while adaptively selecting the optimal encoders for HESP using a multi-armed bandit technique. Our method, MOCA-HESP, is designed as a \textit{meta-algorithm} such that it can incorporate other combinatorial and mixed BO optimizers to further enhance the optimizers' performance. Finally, we develop three practical BO methods by integrating MOCA-HESP with state-of-the-art BO optimizers for combinatorial and mixed variables: standard BO, CASMOPOLITAN, and Bounce. Our experimental results on various synthetic and real-world benchmarks show that our methods outperform existing baselines. 
Our code implementation can be found at \textit{https://github.com/LamNgo1/moca-hesp}.
\end{abstract}

\end{frontmatter}


\newcommand{\bo}{\texttt{BO}\xspace}
\newcommand{\bock}{\texttt{BOCK}\xspace}
\newcommand{\cmaes}{\texttt{CMA-ES}\xspace}
\newcommand{\turbo}{\texttt{TuRBO}\xspace}
\newcommand{\baxus}{\texttt{BAxUS}\xspace}
\newcommand{\lamcts}{\texttt{LA-MCTS}\xspace}
\newcommand{\lamctsbo}{\texttt{LAMCTS-BO}\xspace}
\newcommand{\lamctsturbo}{\texttt{LAMCTS-TuRBO}\xspace}
\newcommand{\mctsvs}{\texttt{MCTS-VS}\xspace}
\newcommand{\mctsvsbo}{\texttt{MCTSVS-BO}\xspace}
\newcommand{\mctsvsturbo}{\texttt{MCTSVS-TuRBO}\xspace}
\newcommand{\dtscmaes}{\texttt{DTS-CMAES}\xspace}
\newcommand{\bads}{\texttt{BADS}\xspace}
\newcommand{\rembo}{\texttt{REMBO}\xspace}
\newcommand{\alebo}{\texttt{ALEBO}\xspace}
\newcommand{\hesbo}{\texttt{HeSBO}\xspace}
\newcommand{\gpop}{\texttt{GPOP}\xspace}
\newcommand{\lqcmaes}{\texttt{lq-CMAES}\xspace}
\newcommand{\casmo}{\texttt{CASMOPOLITAN}\xspace}
\newcommand{\bounce}{\texttt{Bounce}\xspace}
\newcommand{\cmaMeAlg}{\texttt{CMA-Meta-Algorithm}\xspace}
\newcommand{\cmacasmo}{{MOCA-HESP-Casmo}\xspace}
\newcommand{\cmabo}{{MOCA-HESP-BO}\xspace}
\newcommand{\cmabounce}{{MOCA-HESP-Bounce}\xspace}
\newcommand{\cmacma}{{MOCA-HESP}\xspace}
\newcommand{\cocabo}{\texttt{CoCaBO}\xspace}
\newcommand{\smac}{\texttt{SMAC}\xspace}
\newcommand{\tpe}{\texttt{TPE}\xspace}
\newcommand{\combo}{\texttt{COMBO}\xspace}
\newcommand{\bodi}{\texttt{BODi}\xspace}
\newcommand{\ras}{\texttt{Random Search}\xspace}
\newcommand{\bocs}{\texttt{BOCS}\xspace}

\section{Introduction} \label{sec:introduction}
In recent years, high-dimensional BO has emerged as a crucial area of research, as many expensive black-box optimization problems can have hundreds of dimensions. Applications of high-dimensional BO 
include but are not limited to hyperparameter tuning of machine learning models~\cite{Snoek2012PracticalBO,Eriksson2019TuRBO,wang2024pretrainbo}, neural architecture search~\cite{Kandasamy2018NAS, shen2023proxybo}, drug discovery~\cite{khan2022antbo_antibody, guan2022class}, supply chain management~\cite{hu2010contamination_pest, zhang2023forecasting}
and optimal system design~\cite{dreifuerst2021cco}. 

Various research works have tackled the high-dimensional BO problem for continuous spaces~\cite{Eriksson2019TuRBO,papenmeier2022baxus, song2022mctsvs, ziomek2023RDUCB, ngo2024cmabo, ngo2025boids}. However, in many real-world scenarios, objective functions may include combinatorial (ordinal and/or categorical) and mixed variables
\cite{hu2010contamination_pest,ChangYong2019COMBO, dreifuerst2021cco, khan2022antbo_antibody,papenmeier2023bounce}. This presents significant challenges, as BO methods that are tailored for continuous variables cannot be directly applied to solve problems involving these \hybrid variables~\cite{Bin2020CoCaBO,wan2021casmopolitan, papenmeier2023bounce, deshwal2023bodi}. There are multiple reasons.
First, categorical variables lack a natural ordering or rank, preventing standard kernels from capturing the relationships between categorical variables and other variables, thus degrading the surrogate model's predictive accuracy. Second, since ordinal and categorical spaces are discrete, gradient-based optimizers are not effective for optimizing the acquisition function. Third, conventional approaches such as using one-hot encoding to transform categorical variables into numerical values are not scalable as they can significantly increase the problem dimensionality, affecting the computational costs. Although there exist research works that tackle the problem of BO over high-dimensional \hybrid search spaces~\cite{ChangYong2019COMBO,Bin2020CoCaBO,wan2021casmopolitan,daulton2022bopr,papenmeier2023bounce}, due to the aforementioned challenges, achieving promising performance for these types of inputs is still an open question.

In this paper, we propose a novel high-dimensional BO \textit{meta-algorithm} for \hybrid variables. To address high dimensionality, we leverage the hyper-ellipsoid space partitioning technique (HESP)~\cite{ngo2024cmabo} to define local regions with a high probability of containing the global optimum, within which BO can be performed. 
We propose a novel HESP technique designed for combinatorial and mixed problems, which integrates various encoding methods to transform combinatorial variables into continuous representations and employs a multi-armed bandit (MAB) strategy to adaptively select the most effective encoder.
Our proposed method, \emph{\underline{M}eta-Algorithm for High-dimensional \underline{O}rdinal, \underline{CA}tegorical and Mixed Bayesian Optimization via \underline{H}yper-\underline{E}llipsoid \underline{S}pace \underline{P}artitioning} (MOCA-HESP), is designed as a \textit{meta-algorithm}~\cite{Wang2020LAMCTS,song2022mctsvs,ngo2024cmabo} that can incorporate various BO optimizers to solve high-dimensional BO problems with \hybrid variables. 

Furthermore, we develop three novel algorithms \cmabo, \cmacasmo and \cmabounce, corresponding to the cases when we incorporate \cmacma with state-of-the-art BO methods for high-dimensional \hybrid variables: standard BO, CASMOPOLITAN~\cite{wan2021casmopolitan}, and Bounce~\cite{papenmeier2023bounce}, respectively. 
For \cmabo, we integrate \cmacma with the standard BO method, using it as the optimizer within \cmacma. 
For \cmacasmo, we propose a novel technique to integrate CASMOPOLITAN's core feature - the local region adaptation mechanism - into \cmacma and define local regions satisfied both the Mahalanobis and the Hamming distance criteria. 
For \cmabounce, we propose a novel technique to incorporate Bounce's core feature - the subspace embedding technique - into \cmacma's local regions, allowing low-dimensional optimization. 
Notably, the \cmacma meta-algorithm can flexibly adapt different settings, e.g., input data encoding, used by various BO optimizers, making it compatible with existing and future BO optimizers.
Our experimental results on various synthetic and real-world combinatorial and mixed benchmark problems show that the \cmacma methods outperform the respective BO optimizers and other baselines, demonstrating the effectiveness and efficiency of \cmacma. 
We summarize our contributions as follows:
\begin{itemize} 
    \item We propose a novel meta-algorithm, namely \cmacma, for solving BO problems involving high-dimensional \hybrid variables. We develop a novel HESP strategy that leverages categorical encoders and an adaptive encoder selection mechanism to effectively transform combinatorial variables into numerical representations. To the best of our knowledge, \cmacma is the \textit{first meta-algorithm} for BO in high-dimensional \hybrid spaces. 
    \item We develop three practical methods, \cmabo, \cmacasmo, \cmabounce, by proposing novel strategies to incorporate \cmacma with state-of-the-art BO methods for high-dimensional \hybrid variables. 
    \item We conduct extensive experiments and analysis on synthetic and real-world combinatorial and mixed benchmark problems against the baselines, validating the efficiency of our proposed algorithm.
\end{itemize}


\section{Related Work} \label{sec:related-work}
\begin{figure*} [ht]
    \centering
    \includegraphics[width=0.85\linewidth]{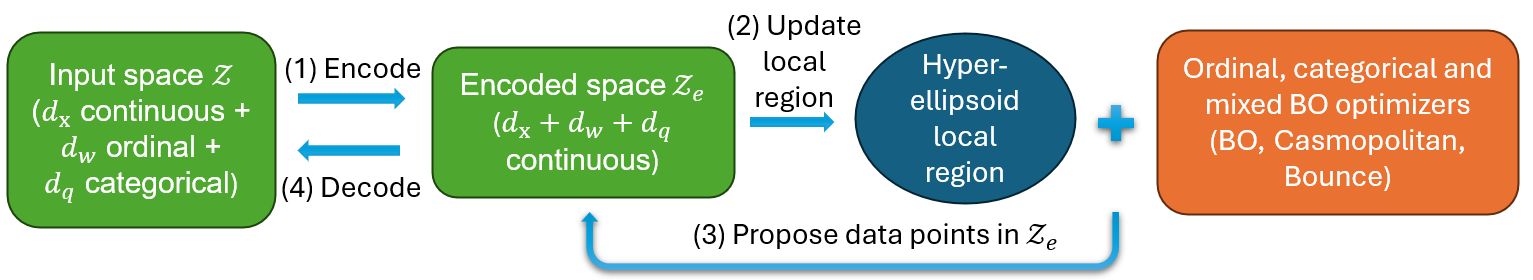}
    \caption{Illustration of \cmacma meta-algorithm. (1) The mixed input data is encoded by an \textit{adaptively chosen encoder}. (2) The hyper-ellipsoid local region, which is derived from the \textit{base local region} (Eq. (\ref{eq:meta_local_region})) and the \textit{distinct features} of the BO optimizers, is constructed/updated, (3) From the hyper-ellipsoid's local region, the BO process (based on the chosen BO optimizes) proposes the next observations in the encoded spaces. (4) A decoder is used to transform the encoded data for function evaluations.}
    \label{fig:illustration}
\end{figure*}

\paragraph{High-dimensional BO with Combinatorial and Mixed Variables.} One popular approach is to enhance the capacity of surrogate models in modeling objective functions with these types of variables. For example, TPE~\cite{bergstra2011TPE} and SMAC~\cite{hutter2011smac} propose to use different surrogate models such as Tree Parzen Structure and Random Forest, which can directly model \hybrid variables. \citeauthor{garrido2020improvedOneHot}~\cite{garrido2020improvedOneHot} propose to modify the kernel by rounding the kernel input to better capture flat regions in \hybrid objective functions. COMBO~\cite{ChangYong2019COMBO} tackles the problem by constructing a combinatorial graph, where each vertex represents a possible combination of categorical variables, and proposing a graph kernel to capture the interactions between categorical variables within this graph. 
\citeauthor{moss2020boss}~\cite{moss2020boss} propose the String kernel, which models the categorical variables as strings, hence performing the optimization in a string space.
CoCaBO~\cite{Bin2020CoCaBO} introduces the Overlapped kernel, which combines continuous and categorical kernels via a weighted average of addition and multiplication to capture the interaction between continuous and categorical variables, enhancing the modeling of the objective function for categorical and mixed variables. Based on this, CASMOPOLITAN~\cite{wan2021casmopolitan} proposes the Transformed Overlapped kernel, improving the expressiveness of the model by learning different GP lengthscales for categorical kernels. BODi~\cite{deshwal2023bodi} employs a dictionary-based embedding to transform combinatorial variables to continuous values, enabling the use of standard continuous GP kernels for modeling \hybrid inputs. However, \citeauthor{papenmeier2023bounce}~\cite{papenmeier2023bounce} showed that BODi's embedding tends to favor generating zero-sequency solutions, which may not be robust in all scenarios. Additionally, BODi does not generalize well to mixed-variable problems beyond binary-continuous settings.

In addition to improving surrogate models, various research works aim to optimize acquisition functions for \hybrid variables more effectively. 
BOCS~\cite{baptista2018bocs} employs Simulated Annealing, which performs random walks in the neighborhood of a data point, moving to the next point based on acquisition function improvements and probabilistic factors. 
COMBO~\cite{ChangYong2019COMBO} employs a local search strategy while applying multi-starts to escape from local optima. 
CoCaBO~\cite{Bin2020CoCaBO} leverages a multi-armed bandit algorithm to optimize the categorical acquisition function. 
CASMOPOLITAN~\cite{wan2021casmopolitan} and Bounce~\cite{papenmeier2023bounce} define local regions to restrict the search space of the acquisition function and then apply a local search in the categorical and mixed space. 
Bounce~\cite{papenmeier2023bounce} further incorporates nested subspace embedding to handle \hybrid spaces more effectively. \citeauthor{daulton2022bopr}~\cite{daulton2022bopr} propose a probabilistic reparameterization to relax the discrete acquisition function optimization into continuous domains, allowing the use of gradient-based optimizers.

\paragraph{The HESP Technique.} The HESP technique~\cite{ngo2024cmabo} is developed from a well-known black-box optimization algorithm in Evolutionary Algorithms (EA), called Covariance Matrix Adaptation Evolutionary Strategy (CMA-ES)~\cite{Hansen2001CMAES}, which aims to address high-dimensional black-box optimization problems. However, CMA-ES is inherently limited to continuous variables. Although several discrete variants have been proposed~\cite{hansen2011cmawIH, hamano2022cmaeswmargin}, none of the CMA-ES based methods are capable of addressing optimization problems involving categorical variables.

\paragraph{Meta-algorithms for BO.} In BO, several meta-algorithms have been developed to incorporate different BO optimizers, enhancing the the optimizers' performance.
LA-MCTS~\cite{Wang2020LAMCTS} learns regions with non-linear boundaries via K-mean classification. 
MCTS-VS~\cite{song2022mctsvs} leverages the Monte Carlo Tree Search to perform dimensionality reduction and select a subset of important variables. The work in~\cite{ngo2024cmabo} defines systematic hyper-ellipsoid local regions using the CMA strategy. However, to the best of our knowledge, all these meta-algorithms only target continuous domains, hence cannot be directly applied to combinatorial and mixed problems.


\section{Background} \label{sec:background}

\subsection{Combinatorial and Mixed Bayesian Optimization} \label{sec:bo-orcatmix}
Let us consider a minimization problem of an \textit{expensive black-box} objective function $f: \mathcal{Z} \rightarrow \mathbb{R}$, where $\mathcal{Z} = \mathcal{X} \times \mathcal{H}$ is a $d$-dimensional mixed domain constructed from a $d_x$-dimensional continuous space $\mathcal{X} \subset \mathbb{R}^{d_x}$, a $d_h$-dimensional combinatorial space $\mathcal{H}$, $d=d_x+d_h$. By definition, $\mathcal H = \mathcal W \times \mathcal Q$ where $\mathcal W$ is a $d_w$-dimensional ordinal space $\mathcal{W}\subset \mathbb{Z}^{d_w}$ and $\mathcal Q$ is a $d_q$-dimensional categorical space $\mathcal{Q}$, $d_h = d_w + d_q$.
Our goal is to find the global minimum $\bm{z}^* = [\bm{x}^*, \bm{h}^*]$ of the function $f$ using \textit{the least number of function evaluations}, 
\begin{equation} \label{eq:problem}
    \bm{z}^* \in \arg{\min_{\bm{z}\in\mathcal{Z}}{f(\bm{z})}},
\end{equation}
where $\bm{z}=[\bm{x}, \bm{h}]$ is the input vector, $\bm{x} \in \mathcal{X}$ is the continuous input vector, $\bm{h} = [\bm{w}, \bm q] \in \mathcal{H}$ is the combinatorial input vector, $\bm{w} \in \mathcal{W}$ is the ordinal input vector, and $\bm{q} = [q_1, ..., q_{d_q}] \in \mathcal{Q}$ is the categorical input vector, with $q_i$ ($i \in [1,..., d_q]$) being a categorical variable having $c_i$ unique categories.
Note that in this work, we treat ordinal variables as categorical variables as in various works~\cite{Bin2020CoCaBO,wan2021casmopolitan, papenmeier2023bounce, deshwal2023bodi}.

In BO, the objective function is assumed to be \textit{expensive}, i.e., time-consuming and/or costly to evaluate, and \textit{black-box}, i.e., the analytical form and other information such as the gradient or Hessian are unavailable.
Therefore, the primary goal of BO is to find the best solution with the least number of function evaluations. 
Details on BO including common surrogate models and acquisition functions for \hybrid BO are in the 
Appendix~\ref{sec:appendix_bo} and~\ref{sec:appendix_bo_model_acq}.

\subsection{The Hyper-ellipsoid Space Partitioning Strategy}
The HESP technique is a search space partitioning method designed to solve \textit{high-dimensional} BO problems for continuous variables~\cite{ngo2024cmabo}. The key idea is to maintain a \textit{multivariate normal search distribution} $\mathcal{N}(\mathbf{m},\mathbf{\Sigma})$ over the input search space, where $\mathbf{m}$ is the mean vector and $\mathbf{\Sigma}$ is the covariance matrix of the search distribution. A local region is then defined as the $\alpha$-confidence hyper-ellipsoid of $\mathcal{N}(\mathbf{m},\mathbf{\Sigma})$. Within the local region, a BO process can be used to sequentially suggest $\lambda > 1$ data points for function evaluation. These $\lambda$ observations are subsequently used to update the search distribution $\mathcal{N}(\mathbf{m},\mathbf{\Sigma})$ and the local region in the next iteration via the mean update and covariance matrix adaptation formulas~\cite{Hansen2001CMAES}. The local regions defined by HESP are shown to have a high probability of containing the global optimum of the objective function. The HESP strategy has been shown to not only perform well in finding  global optima of high-dimensional optimization problems but also incurs minimal computation overhead~\cite{ngo2024cmabo}. However, despite its effectiveness, HESP can only work for continuous variables.


\section{Proposed Methods} \label{sec:method}

In this section, we first develop a HESP strategy tailored for \hybrid variables (Sec.~\ref{sec:cma_for_mixed}). Then, we present our proposed meta-algorithm, \cmacma, for BO in high-dimensional \hybrid spaces (Sec.~\ref{sec:camicma}). Finally, we develop three algorithms: \cmabo, \cmacasmo and \cmabounce, obtained when incorporating \cmacma with the state-of-the-art high-dimensional BO methods for these mixed spaces: standard BO, CASMOPOLITAN~\cite{wan2021casmopolitan}, and Bounce~\cite{papenmeier2023bounce} (Sec.~\ref{sec:camicma_derived}). Incorporating \cmacma with CASMOPOLITAN and Bounce is especially challenging due to their distinct core features, such as different local region strategies or nested subspace embedding mechanisms, as well as their different requirements such as input data encoding. For an overview of these optimizers, see the 
Appendix~\ref{sec:appendix_casmo} and~\ref{sec:appendix_bounce}.

\subsection{The Proposed HESP Strategy for Combinatorial and Mixed Spaces} \label{sec:cma_for_mixed}
To develop HESP for \hybrid variables, we propose to use categorical encoders~\cite{potdar2017CategoricalEncoding} to transform combinatorial (ordinal and categorical) variables into continuous values and apply HESP to these transformed variables. In addition, we also suggest a decoder to transform the encoded data back to the original domain for objective function evaluations. We further propose to adaptively select the optimal encoder for HESP using a multi-armed bandit (MAB) algorithm. Below, we describe these steps in detail.

\paragraph{Combinatorial and Mixed Variables Encoding and Decoding for HESP.}
From the mixed space $\mathcal Z=\mathcal X \times \mathcal H$, we convert combinatorial variables to continuous values by using a 1-to-1 mapping categorical encoders $g: \mathcal{H} \rightarrow \mathcal{H}_e$ where $\mathcal{H}$ is a $d_h$-dimensional combinatorial space and $\mathcal{H}_e \subset \mathbb{R}^{d_h}$. In this work, we propose to use either an \textit{ordinal} or a \textit{target} encoder~\cite{micci2001targetencoder} as these encoders are commonly used, effective, efficient in runtime, and do not require as much training data as a VAE~\cite{kingma2019vae}. Ordinal encoders map the inputs to numerical values based on existing or assumed relationships on their orders and rankings. Target encoders map the inputs to numerical values based on their target values, e.g., the average function values. More details on these two encoders are in the 
Appendix~\ref{sec:appendix_categorical_encoder}. 
After training the encoder $g(.)$, we apply the HESP technique to the encoded mixed spaces $\mathcal Z_e = \mathcal X \times \mathcal H_e$. Note that, in order to perform objective function evaluation on $\mathcal Z$, we need a decoder $\tilde g: \mathcal H_e \rightarrow \mathcal H$ to transform the encoded  continuous values in $\mathcal{H}_e$ back to the original combinatorial domain $\mathcal{H}$. In particular, to decode a numerical value $x\in\mathcal{H}_e$ to a $c$-choice combinatorial variable $h \in \{u_1,\dots,u_c\}$, we find a category $u_i$ with the encoded value closest  to the numerical value $x$, or formally $\tilde{g}(x) = u_i$ where $i = \arg{\min_{i= 1,\dots,c}{|g(u_i)-x|}}$.

\paragraph{Adaptive Encoder Selection.} In practice, as no encoder is guaranteed to be the best in all cases, to select the optimal encoders for HESP, we propose to use a multi-armed bandit (MAB) approach. We consider each encoder as an MAB action, and the goal is to select the best action (i.e., the best encoder), based on the optimization data. We propose to use EXP3~\cite{auer2002exp3} as the MAB approach due to its general applicability, robustness to adversarial environments, and strong theoretical regret guarantees.
Our proposed adaptive encoder selection method is outlined in the pseudo-code Alg.~\ref{alg:exp3_cma}. Each encoder corresponds to an action, and is associated with a weight and a selection probability, which determines the likelihood of that action being chosen. Given $K$ types of encoders and their associated weights $w_k(t)$ for $k=1,2,\dots,K$ at iteration $t$, the selection probability for each action $k$ (line~\ref{alg-line:compute_prob}) is:
\begin{equation} \label{eq:exp3}
    p_k(t) = (1-\eta)\frac{w_k(t)}{\textstyle\sum_{j=1}^{K}{w_j(t)}} + \frac{\eta}{K},
\end{equation}
where $\eta \in [0,1]$ is a parameter that governs how much we should sample the actions uniformly at random (explore unknown actions) rather than focus on good performing actions (exploit potential actions)~\cite{auer2002exp3}. In each iteration, to update the weights $w_k(t)$, we compute the estimated reward $\hat{r}(t)$ from the $\lambda$ observations in an iteration of \cmacma.
We first normalize the observed function values ${y}_i|_{i=1}^{t}$ to $\hat{y}_i|_{i=1}^{t}$ such that $\hat{y}_i = (y_i - \max{\bm{y}}) / (\min{\bm{y}} - \max{\bm{y}})$ (line~\ref{alg-line:min-max_normalize}). This normalization ensures that a high reward $\hat{r}(t)$ is given only when the reward $\hat{y}_i$ is similarly good or better, increasing the action’s weight and selection probability~\cite{ahmadianshalchi2024pareto}. Otherwise, the probability remains unchanged, allowing EXP3 to explore other actions.
The estimated reward is then calculated as $\hat{r}(t) = r(t)/p_{k_{t}}(t)$, where $r(t) = \min{\hat{y}_i|_{k=t-\lambda+1}^{t}}$ is the minimum normalized function value (lines~\ref{alg-line:select_reward}-\ref{alg-line:estimated_reward}).
Subsequently, we update the current selected action's weight $w_{k_{t}}(t+1) = w_{k_t}(t)\exp(\eta \hat{r}(t) / K)$, and preserve other actions' weights, $w_j(t+1) = w_j(t)$, $j\neq k_t$ (line~\ref{alg-line:update_exp3_weight}). 
Finally, a new selection probability is computed (line~\ref{alg-line:update_exp3_prob}) to sample a new encoder type $k_{t+1}$ for the next iteration (line~\ref{alg-line:sample_new_encoder}). 
 
\begin{algorithm}[ht] 
   \caption{The Adaptive Encoder Selection.}
   \label{alg:exp3_cma}
\begin{algorithmic}[1]
   \State {\bfseries Input:} The observed function values $y_j|_{j=1}^{t}$ up to iteration $t$, the number of categorical encoder types $K$, the weight vector $w_k(t)|_{k=1}^{K}$, the current encoder type $k_t$, the parameters $\eta$ and $\lambda$
   \State {\bfseries Output:} The next encoder type $k_{t+1}$.
   \State Compute the probability $p_k(t)$, $k=1,\dots,K$ \Comment{Eq. (\ref{eq:exp3})} \label{alg-line:compute_prob}
   \State Normalize $y_i|_{i=1}^{n}$ to $\hat{y}_i|_{i=1}^{t}$, where $\hat{y}_i\in [0,1]$ \label{alg-line:min-max_normalize}
   \State Compute the reward $r(t) = \min{\hat{y}_i|_{k=t-\lambda+1}^{t}}$. \label{alg-line:select_reward}
   \State Compute the estimated reward $\hat{r}(t) = r(t)/p_{k_t}(t)$ \label{alg-line:estimated_reward}
   \State Update the weights $w_{k}(t+1)$, $k=1,\dots,K$  \label{alg-line:update_exp3_weight}
   \State Update the selection probability $p_k(t+1)$, $k=1,\dots,K$ \label{alg-line:update_exp3_prob}
   \State Sample the next encoder type $k_{t+1}$ following $p_k(t+1)$ \label{alg-line:sample_new_encoder}
   \State Return the encoder type $k_{t+1}$
\end{algorithmic}
\end{algorithm}

\begin{algorithm}[tb] 
   \caption{The \cmacma Meta-algorithm.}
   \label{alg:cami-cma}
\begin{algorithmic}[1]
   \State {\bfseries Input:} The objective function $f(.)$, the mixed search domain $\mathcal{Z}$, the evaluation budget $N$, the number of initial points $n_0$, the BO optimizer $\texttt{bo\_opt}$, the number of categorical encoder types $K$
   \State {\bfseries Output:} The optimum $\bm{z}^*$
   \State Set $\lambda$, $t\leftarrow 0$, $T\leftarrow \lfloor(N-n_0)/\lambda\rfloor$, the dataset $\mathcal{D}\leftarrow\emptyset$
   \While {$t \le T$}
   \State Set the encoder weights $w_k=1$, for $i=1\dots K$
   \State Choose and train the categorical encoder $g_k^{(t)}(.)$ \Comment {Eq. (\ref{eq:exp3})}\label{alg-line:choose_encoder_uniformly_random} 
   \State Sample a set of $n_0$ initial data points $\mathcal{D}_0$ \label{alg-line:init_data}
   \State Set the initial HE local region $\mathcal{E}^{(0)}_{\texttt{bo\_opt}}$ based on $g_k^{(t)}(\mathcal{D}_0)$\label{alg-line:init_distr}
   \State Set $\mathcal{D} \leftarrow \mathcal{D}\cup \mathcal{D}_0$
   \While{$t \le T$}
   \State Use $\texttt{bo\_opt}$ to propose a set of $\lambda$ data points $\mathcal{D}_\lambda \in \mathcal{Z}$. \label{alg-line:apply_bo_opt}
   \State Update $\mathcal{E}^{(t+1)}_{\texttt{bo\_opt}} \leftarrow \Call{HESP}{g_k^{(t)}(\mathcal{D}_\lambda)}$ \label{alg-line:cma_update} 
   \State Update $\mathcal{D} \leftarrow \mathcal{D}\cup \mathcal{D}_\lambda$, $t \leftarrow t + 1$
   \State Select and train a new encoder $g_k^{(t+1)}(.)$ \label{alg-line:pick_new_encoder} \Comment{Alg.~\ref{alg:exp3_cma}}
   \State Break loop {\bfseries if} the stopping criteria are satisfied \label{alg-line:restart}
   \EndWhile
   \EndWhile
   \State Return $\bm{z}^* = \arg\min_{\bm{z}_i \in \mathcal{D}} \{y_i\}_{i=1}^N$ from the dataset $\mathcal{D}$
\end{algorithmic}
\end{algorithm}

\subsection{The Proposed \cmacma Meta-Algorithm} \label{sec:camicma}

In this section, we present our proposed meta-algorithm \cmacma for BO in high-dimensional \hybrid spaces.

\paragraph{Overall Process.}

The overall process of \cmacma is shown 
in Alg.~\ref{alg:cami-cma} and illustrated in Fig.~\ref{fig:illustration}. Initially, all $K$ encoder weights are set as 1 (line~\ref{alg-line:choose_encoder_uniformly_random}). Based on the initial observed dataset $\mathcal{D}_0 \in \mathcal{Z}$ (line~\ref{alg-line:init_data}), we construct an initial hyper-ellipsoid local region $\mathcal{E}^{(0)}_{\texttt{bo\_opt}}$ in the encoded space $\mathcal{Z}_e \subseteq \mathbb{R}^{d}$ (line~\ref{alg-line:init_distr}). 
Subsequently, within the local region $\mathcal{E}^{(0)}_{\texttt{bo\_opt}}$, the BO optimizer  $\texttt{bo\_opt}$ suggests $\lambda$ data points $\mathcal{D}_\lambda \in \mathcal{Z}$ (line~\ref{alg-line:apply_bo_opt}). Then, the encoded $\lambda$ data points $g_k^{(0)}(\mathcal{D}_\lambda)$ are used to update the local region $\mathcal{E}^{(1)}_{\texttt{bo\_opt}}$ (line~\ref{alg-line:cma_update}). Furthermore, a new categorical encoder $g_k^{(1)}(.)$ is also suggested for the next iteration based on Alg.~\ref{alg:exp3_cma} (line~\ref{alg-line:pick_new_encoder}). Restart conditions~\cite{Hansen2016CMAESTutorial, ngo2024cmabo} are verified every iteration to prevent the algorithm from being trapped at a local optimum (line~\ref{alg-line:restart}). This process repeats until the evaluation budget is depleted, returning the best data point in dataset $\mathcal{D}$ as the optimum $\bm{z}^*$. In the below we describe in detail each component of our proposed \cmacma meta-algorithm.

\paragraph{Local Region Formulation.} 
We provide a detailed formulation of the local regions for HESP in \hybrid domains given the aforementioned categorical encoders.
We first define a \textit{base local region} $\mathcal{E}_{b}$ based on a \textit{multivariate normal search distribution} $\mathcal{N}_{\mathcal{Z}_e}=\mathcal{N}_{\mathcal{Z}_e} (\bm{m}_{\mathcal{Z}_e},\mathbf{\Sigma}_{\mathcal{Z}_e})$ in the encoded space $\mathcal{Z}_e$, where $\bm{m}_{\mathcal{Z}_e}$ and $\mathbf{\Sigma}_{\mathcal{Z}_e}$ are the encoded mean vector and covariance matrix of the search distribution.
Specifically, $\mathcal{E}_{b}$ is defined as the $\alpha$-confidence hyper-ellipsoid of the search distribution, i.e.,
\begin{equation} \label{eq:meta_local_region}
    \mathcal{E}_{b} = \left\{\bm{z}_e \mid \mathbb{M}\left(\bm{z}_e, \mathcal{N}_{\mathcal{Z}_e}\right) \leq \mathbf\chi^2_{1-\alpha}(d_{\mathcal{Z}_e})\right\},
\end{equation}
where $\mathbb{M}(\bm{x}, \mathcal{N})=\sqrt{(\bm{x} - {\bm{m}})^{\intercal}\mathbf{\Sigma}(\bm{x} - {\bm{m}})}$ is the Mahalanobis distance from a continuous vector $\bm{x}$ to a multivariate normal distribution $\mathcal{N}=\mathcal{N}(\bm{m},\mathbf{\Sigma})$ and $\chi^2_{1-\beta}(\gamma)$ denotes a Chi-squared $\beta$ critical value with $\gamma$ degree of freedom. From the base local region $\mathcal{E}_{b}$, additional requirements from the respective BO optimizers will be incorporated to create the local region $\mathcal{E}_{\texttt{bo\_opt}}$ in later sections.

\paragraph{Local Optimization.}
In each iteration, we apply the BO optimizer \texttt{bo\_opt} within the local region $\mathcal{E}_{\texttt{bo\_opt}}$ to propose $\lambda$ data points in $\bm{z} \in \mathcal{Z}$. When fitting the GP, we preserve all settings from the BO optimizer, e.g., the kernel and input encoding type. To propose the next observed data, we optimize the same acquisition function employed by the BO optimizer. Specifically, we first sample a pool of candidate data points ${\bm{z}_e}_\alpha\in \mathcal{Z}_e$ following $\mathcal{N}_{\mathcal{Z}_e}(\bm{m}_{\mathcal{Z}_e},\mathbf{\Sigma}_{\mathcal{Z}_e})$ and confined within $\mathcal{E}_{\texttt{bo\_opt}}$. These candidate points assist the acquisition function optimization depending on the type of acquisition function employed by the BO optimizer. Furthermore, the next observation data points must satisfy the local region $\mathcal{E}_{\texttt{bo\_opt}}$. Details for each optimizer will be explained in the next sections. Note that if the acquisition functions are defined in the original mixed space $\mathcal{Z}$, we use a decoder to transform ${\bm{z}_e}_\alpha$ to $z_\alpha \in \mathcal{Z}$ to compute the acquisition function values.

\subsection{The Proposed \cmabo, \cmacasmo, \cmabounce} \label{sec:camicma_derived}

\begin{table*}[ht]
\centering
\caption{Key ideas in the three \cmacma algorithms. \cmacma can robustly incorporate different BO optimizers that (1)  require different settings, e.g., data encoding, kernel, acquisition function, and (2) have distinct core features, e.g., local region adaptation mechanisms, subspace embedding techniques.}
\begin{tabular}{|l|p{4.3cm}|p{9.7cm}|}
\hline
\textbf{\cmacma Method} & \textbf{Local Region} & \textbf{Local Optimization} \\
\hline
\cmabo & 
\begin{minipage}[t]{\linewidth}
\begin{itemize}[noitemsep, topsep=10pt, parsep=0pt, partopsep=0pt,leftmargin=*, nosep]
    \item In the encoded space $\mathcal Z_e$.
    \item Local regions $\mathcal{E}_\text{BO}$ satisfy the Mahalanobis distance criterion.
\end{itemize}
\end{minipage}
&
\begin{minipage}[t]{\linewidth}
\begin{itemize}[noitemsep, topsep=-\parskip, parsep=0pt, partopsep=0pt,leftmargin=*, nosep]
    \item Train a GP with ordinal encoded data (Matern 5/2 kernel).
    \item Sample candidates in $\mathcal Z_e$ via a scaled search distribution (for Mahalanobis criterion).
    \item Decode candidates back to $\mathcal Z$.
    \item Maximize the acquisition function over candidates.
\end{itemize}
\end{minipage}
\\
\hline
\cmacasmo & 
\begin{minipage}[t]{\linewidth}
\begin{itemize}[noitemsep, topsep=-\parskip, parsep=0pt, partopsep=0pt,leftmargin=*, nosep]
    \item In the encoded space $\mathcal Z_e$.
    \item Local regions $\mathcal{E}_\text{Casmo}$ satisfy the Mahalanobis distance criterion \textit{and} the Hamming distance criterion.
\end{itemize}
\end{minipage}
&
\begin{minipage}[t]{\linewidth}
\begin{itemize}[noitemsep, topsep=-\parskip, parsep=0pt, partopsep=0pt,leftmargin=*, nosep]
    \item Train a GP with ordinal encoded data (Transform Overlapped kernel).
    \item Sample candidates in $\mathcal Z_e$ via a scaled search distribution (for Mahalanobis criterion).
    \item Decode candidates back to $\mathcal Z$.
    \item Randomly reset several categorical variables to satisfy the Hamming criterion.
    \item Maximize the acquisition function over candidates.
\end{itemize}
\end{minipage}
\\
\hline
\cmabounce & 
\begin{minipage}[t]{\linewidth}
\begin{itemize}[noitemsep, topsep=-\parskip, parsep=0pt, partopsep=0pt,leftmargin=*, nosep]
    \item In the encoded subspace $\mathcal V$.
    \item Local regions $\mathcal{E}_\text{Bounce}$ satisfy the Mahalanobis distance criterion \textit{and} the Hamming distance criterion.
\end{itemize}
\end{minipage}
&
\begin{minipage}[t]{\linewidth}
\begin{itemize}
    \item Train a GP with one-hot encoded data (Matern 5/2 kernel).
    \item Sample candidates in $\mathcal V$ via a scaled search distribution (for Mahalanobis criterion).
    \item Decode candidates back to the subspace $\mathcal A$.
    \item Randomly reset several categorical variables to satisfy the Hamming criterion.
    \item Maximize the acquisition function in $\mathcal A$ and project the best candidate to $\mathcal Z$:
    \begin{itemize}[label=--,noitemsep, nosep]
        \item Categorical problems: Local search over candidates.
        \item Mixed problems: Interleave search the categorical and continuous variables separately. While optimizing categorical variables, fix the candidates' continuous variables to the current best solution, and vice versa.
    \end{itemize}
\end{itemize}
\end{minipage}
\vspace{0.01mm}
\\
\hline
\end{tabular}
\label{table:method_summary}
\end{table*}

In this section, we develop three algorithms by incorporating our meta-algorithm \cmacma with the state-of-the-art \hybrid BO optimizers: standard BO, CASMOPOLITAN~\cite{wan2021casmopolitan}, and Bounce~\cite{papenmeier2023bounce}. Incorporating \cmacma with CASMOPOLITAN and Bounce is especially challenging, as these optimizers have \textit{different requirements} for input data encoding, kernel type, and acquisition function, as well as \textit{distinct components} such as local region formulations and subspace embedding techniques. The key ideas of the three proposed methods are summarized in Table~\ref{table:method_summary}.

\subsubsection{The \cmabo Algorithm} \label{sec:cmabo}
This algorithm is developed by incorporating \cmacma with the standard BO optimizer. \cmabo shares similar components with the meta-algorithm \cmacma, including the formulation of the local region $\mathcal{E}_\text{BO}$ and the local optimization step.

\subsubsection{The \cmacasmo Algorithm} \label{sec:cmacasmo}
This algorithm is developed by incorporating our meta-algorithm \cmacma with CASMOPOLITAN~\cite{wan2021casmopolitan}. The challenge is that we need to incorporate CASMOPOLITAN's local region adaptation mechanism, which maintains two separate local regions for continuous and categorical variables. 
To overcome this challenge, first, we incorporate the continuous local region (governed by a scaling factor $L_x$) by scaling the base local region in Eq. (\ref{eq:meta_local_region}) by $L_x$. In particular, we scale the covariance matrix $\mathbf{\Sigma}_{\mathcal{Z}_e}$ by a $d_{\mathcal{Z}_e}$-dimensional scaling vector $\bm{L}$, where $L_i = L_x^2$ if the $i$-th dimension is associated with a continuous variable, and $L_i = 1$ otherwise. This results in a scaled distribution $\mathcal{N}_{\mathcal{Z}_e,\bm{L}}=\mathcal{N}_{\mathcal{Z}_e}(\bm{m}_{\mathcal{Z}_e}, \psi(\mathbf{\Sigma}_{\mathcal{Z}_e}, \bm{L}))$, where $\psi(\mathbf{\Sigma}, \bm{S})$ scales the radii of the $\alpha$-level confidence hyper-ellipsoid of $\mathbf{\Sigma}$ by the vector $\bm{S}$. 
Second, to incorporate the categorical local region (governed by a Hamming distance threshold $L_h$), we impose an additional threshold to the base local region. That is, we employ additional constraint $\mathbb{H}\left(\tilde{g}_k(\bm{z}_e), \tilde{g}_k(\bm{m}_{\mathcal{Z}_e})\right) \leq L_h$, where $\mathbb{H}(.,.)$ is the Hamming distance between categorical data points, and $\tilde{g}_k(.)$ is the categorical decoder from $\mathcal{Z}_e$ to $\mathcal{Z}$. Overall, the local region $\mathcal{E}_{Casmo}$ is the combination of two requirements, such that,
\begin{equation} \label{eq:casmo_local_region}
\begin{split}
    \mathcal{E}_\text{Casmo} = \{\bm{z}_e \mid &\mathbb{M}\left(\bm{z}_e,\mathcal{N}_{\mathcal{Z}_e,\bm{L}}\right) \leq \mathbf\chi^2_{1-\alpha}(d_{\mathcal{Z}_e}); 
    \\&\mathbb{H}\left(\tilde{g}_k(\bm{z}_e), \tilde{g}_k(\bm{m}_{\mathcal{Z}_e})\right) \leq L_h\}.
\end{split}
\end{equation}
Having defined the local region $\mathcal{E}_\text{Casmo}$, we carry out the local optimization to select the next observation data point. The choice of the surrogate model and the acquisition function are similar to CASMOPOLITAN, however, acquisition optimization needs to satisfy the local region $\mathcal{E}_\text{Casmo}$. Specifically, we sample a pool of candidates $\mathbf{z}_e$ from the scaled search distribution $\mathcal{N}_{\mathcal{Z}_e,\bm{L}}$ that satisfy the Mahalanobis distance criterion. Then we decode these candidates back to the mixed spaces $\mathcal Z$ via the decoder $\tilde{g}_k(.)$. Then, random selection is carried out to alter several dimensions of $\tilde{g}_k(\mathbf{z}_e)$ until they all satisfy the Hamming distance criterion. New observations are chosen from these candidates by maximizing the acquisition function values.

\subsubsection{The \cmabounce Algorithm} 
\label{sec:cmabounce}
This algorithm is developed by incorporating \cmacma with Bounce~\cite{papenmeier2023bounce}, a recent state-of-the-art BO method for combinatorial and mixed variables.  
The challenge is how to incorporate Bounce's expanding subspace embedding technique, which optimizes a series of mixed low-dimensional subspaces $\mathcal{A}$ (\textit{mixed target subspace}) of increasing dimensions $d_\mathcal{A}$. For an overview of subspace embedding in Bounce, refer to the 
Appendix~\ref{sec:appendix_bounce}.
We denote the \textit{encoded target subspace} $\mathcal{V} \subset \mathbb{R}^{d_\mathcal{A}}$ as a $d_\mathcal{V}$-dimensional encoded space of $\mathcal{A}$.
The relationship between these spaces is given by $d_\mathcal{A} = d_\mathcal{V} < d_{\mathcal{Z}_e} = d_\mathcal{Z}$. In each iteration of \cmabounce, we first derive the base local region (Eq. (\ref{eq:meta_local_region})) in the encoded space $\mathcal{V}$ via a projected search distribution $\mathcal{N}_\mathcal{V}(\bm{m}_\mathcal{V}, \mathbf{\Sigma}_\mathcal{V})$, where $\bm{m}_\mathcal{V}$, $\mathbf{\Sigma}_\mathcal{V}$ are the mean vector and the covariance matrix, respectively. To compute $\mathcal{N}_\mathcal{V}(\bm{m}_\mathcal{V}, \mathbf{\Sigma}_\mathcal{V})$, we project the search distribution $\mathcal{N}_{\mathcal{Z}_e}(\bm{m}_{\mathcal{Z}_e},\mathbf{\Sigma}_{\mathcal{Z}_e})$ onto $\mathcal{V}$ using the embedding matrix $\bm{P}: \mathcal{Z}_e \rightarrow \mathcal{V}$, which is the Moore–Penrose inverse of Bounce's projection matrix $\bm{Q}: \mathcal{V} \rightarrow \mathcal{Z}_e$~\cite{golub2013matrix, ngo2024cmabo}. 
Subsequently, as Bounce employs CASMOPOLITAN's local region adaptation mechanism, we apply the similar technique as in \cmacasmo to define the respective local region in $\mathcal{V}$ - via scaling the base local region by a continuous factor $L_x$ and imposing a Hamming constraint by a categorical factor $L_h$.
Hence, the local region $\mathcal{E}_{Bounce}$ is,
\begin{equation} \label{eq:bounce_local_region}
\begin{split}
    \mathcal{E}_\text{Bounce} = \{\bm{v} \mid &\mathbb{M}\left(\bm{v},\mathcal{N}_{\mathcal{V}, \bm{L}}\right) \leq \mathbf\chi^2_{1-\alpha}({d_{\mathcal{V}}}); 
    \\&\mathbb{H}(\tilde{h}_k(\bm{v}), \tilde{h}_k(\bm{m}_{\mathcal{V}})) \leq L_h\},
\end{split}
\end{equation}
where $\mathcal{N}_{\mathcal{V}, \bm{L}}=\mathcal{N}_{\mathcal{V}, \bm{L}}(\bm{m}_\mathcal{V}, \psi(\mathbf{\Sigma}_{\mathcal{V}}, \bm{L}))$ and $\tilde{h}_k(.)$ is the categorical decoder.
Having defined the local region $\mathcal{E}_\text{Bounce}$, we carry out the local optimization to select the next observation data point. The choice of the surrogate model and the acquisition function are similar to Bounce, yet the acquisition optimization needs to satisfy  $\mathcal{E}_\text{Bounce}$. Specifically, the candidate sampling steps need to satisfy both the Mahalanobis distance and the Hamming distance criterion in the encoded target subspace $\mathcal V$. For categorical problems, we use local search over the generated candidates. For mixed problems, we leverage the interleave search procedure and optimize combinatorial and continuous parts separately. While optimizing combinatorial variables, we fix the continuous values of the candidates, and vice versa. 
Note that as \cmabounce suggests data points in $\mathcal{V}$, we decode the solution to $\mathcal{A}$, then project it to $\mathcal{Z}$ for function evaluations.

\begin{figure*} [ht]
  \centering
    \includegraphics[trim={0 0.5cm 0 0cm}, width=0.99\textwidth]{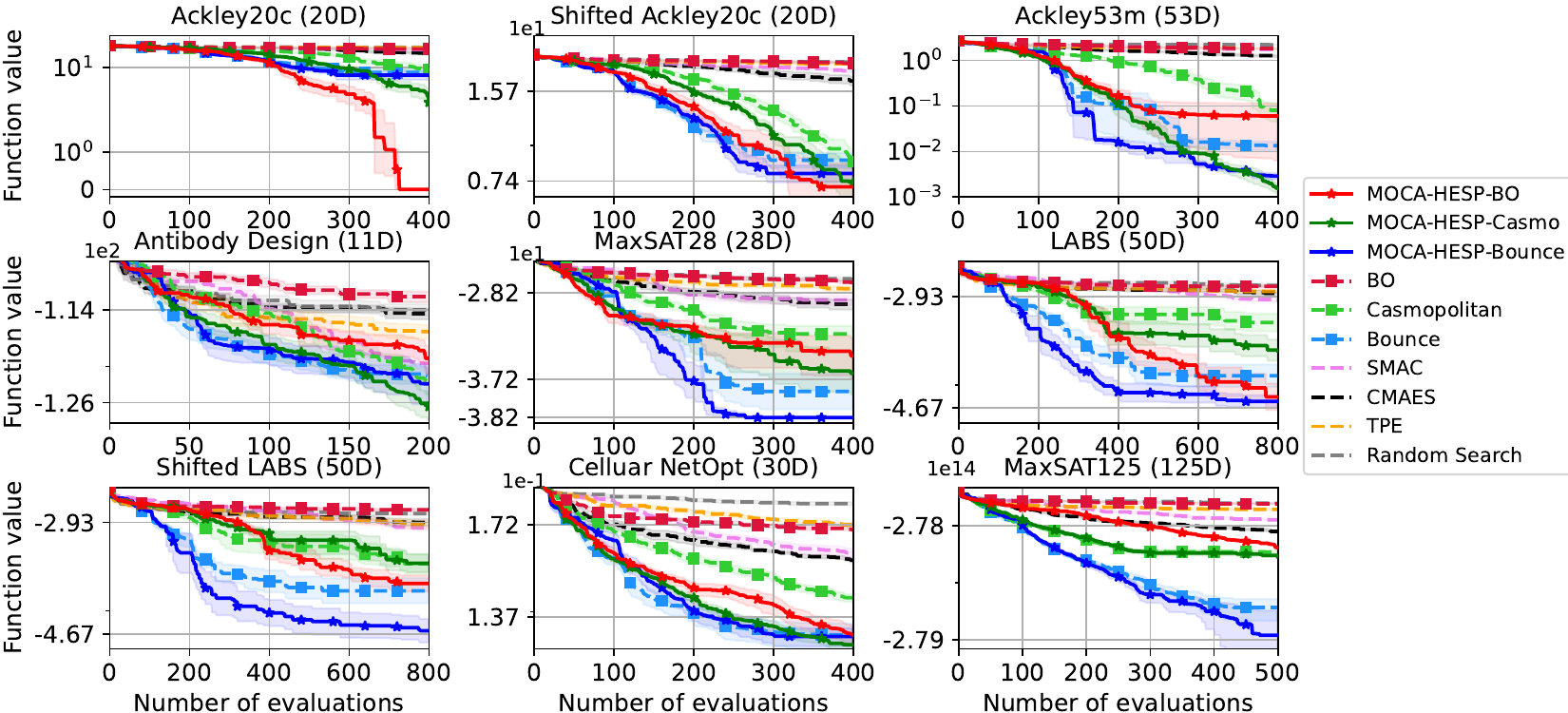}
\caption{Comparison between the \cmacma methods (\cmabo, \cmacasmo, \cmabounce) 
against (1) the original BO optimizers (BO, CASMOPOLITAN, Bounce) respectively and (2) the other baselines (TPE, SMAC, CMA-ES, Random Search). The results show that all the \cmacma methods outperform or are on par with their respective BO optimizers (e.g. \cmabo outperforms BO, \cmacasmo outperforms CASMOPOLITAN, and \cmabounce outperforms Bounce), and also outperform the other baselines.} \label{fig:results}
\end{figure*}


\section{Experiments} \label{sec:experiment}

\subsection{Experiment Setup and Baselines}
We extensively evaluate our proposed \cmacma meta-algorithm by comparing the three developed algorithms \cmabo, \cmacasmo and \cmabounce against: (1) their respective BO optimizers: \textbf{Standard BO}, \textbf{CASMOPOLITAN}~\cite{wan2021casmopolitan}, \textbf{Bounce}~\cite{papenmeier2023bounce}, (2) other state-of-the-art BO optimizers for high-dimensional \hybrid variables including \textbf{TPE}~\cite{bergstra2011TPE}, \textbf{SMAC}~\cite{hutter2011smac}, (3) \textbf{CMA-ES}~\cite{Hansen2001CMAES}, an evolutionary optimization algorithm, in which we used the ordinal encoder to transform combinatorial inputs to numeric values, and (4) \textbf{Random Search}. Note that there are no meta-algorithm baselines as to the best of our knowledge, we are the first to propose a meta-algorithm for high-dimensional BO in combinatorial and mixed spaces.
Details of the settings for \cmacma and the baselines are in the 
Appendix~\ref{appendix:experiment_setup} and Appendix~\ref{appendix:baseline_details}.

\subsection{Benchmark Problems}
We conduct experiments on 9 benchmark problems: 3 synthetic and 6 real-world problems. 
The synthetic problems include 2 combinatorial problems: \textit{Ackley20c - 20D} and \textit{Shifted Ackley20c - 20D}, and 1 mixed problem: \textit{Ackley53m - 53D}~\cite{Bin2020CoCaBO, wan2021casmopolitan, papenmeier2023bounce, deshwal2023bodi}.
The real-world problems include 5 combinatorial ones: \textit{Antibody Design - 11D}~\cite{khan2022antbo_antibody}, \textit{LABS - 50D}~\cite{packebusch2016labs}, \textit{MaxSAT28 - 28D}, \textit{MaxSAT125-125D}~\cite{hansen1990maxsat}, \textit{Shifted LABS - 50D}, and 1 mixed problem: \textit{Cellular Network - 30D}~\cite{dreifuerst2021cco}. The dimensionalities of these problems range from \textbf{11 to 125}, and the cardinalities of the combinatorial variables range from \textbf{2 to 20}. 
These problems have been widely used in other works to evaluate high-dimensional BO methods for \hybrid spaces~\cite{ChangYong2019COMBO,Bin2020CoCaBO,wan2021casmopolitan,daulton2022bopr,deshwal2023bodi}. Note that we also evaluate our methods using shifted benchmark functions, which are designed to eliminate the special structure, e.g., zero-sequency patterns, of the global optimum in the original test functions~\cite{papenmeier2023bounce, ngo2024cmabo}.
Details of these benchmark problems are in Table~\ref{table:test_problems} and the 
Appendix~\ref{appendix:experiment_details}.

\subsection{Performance Comparison with Baselines}

\begin{table} [ht]
\caption{Details of benchmark problems.}
\begin{center}
\begin{tabular}{|l|l|c|}
\hline
\textbf{\makecell[l]{Benchmark\\problems}} & \textbf{\makecell[l]{Inputs ($\bm{x}$ is continuous,\\$\bm{h}$ is combinatorial)}} & \textbf{\makecell[l]{\# evals}} \\
\hline
\makecell[l]{Ackley20c\\(20D)}
& \makecell[l]{$\bm{h} \in [-32.768,..., 32.768] ^{20}$,\\each has 11 evenly spaced values}& 400\\ 
\hline
\makecell[l]{Antibody Design\\(11D)}
& \makecell[l]{$\bm{h} \in AA ^{11}$,\\where AA is the set of 20 amino acids\\ } 
& 200\\
\hline
\makecell[l]{LABS\\(50D)}
& \makecell[l]{$\bm{h} \in \{-1, +1\} ^{50}$} 
& 800\\
\hline
\makecell[l]{MaxSAT28\\(28D)}
& \makecell[l]{$\bm{h} \in \{0, 1\} ^{28}$} 
& 400\\
\hline
\makecell[l]{Ackley53m\\(53D)}
& \makecell[l]{$\bm{x} \in [-1,1] ^{3}$ \\
$\bm{h} \in \{0,1\} ^{50}$} 
& 400\\
\hline
\makecell[l]{Cellular Network\\(30D)}
& \makecell[l]{$\bm{x} \in [30, 50] ^{15}$ \\ $\bm{h} \in \{0, 5\} ^{15}$} 
& 400\\
\hline
\makecell[l]{MaxSAT125\\(125D)}
& \makecell[l]{$\bm{h} \in \{0, 1\} ^{25}$} 
& 500\\
\hline
\end{tabular}
\end{center}
\label{table:test_problems}
\end{table}

\begin{figure*} [t] 
\centering  
  \includegraphics[trim={0 0.5cm 0 0cm}, width=0.97\textwidth]{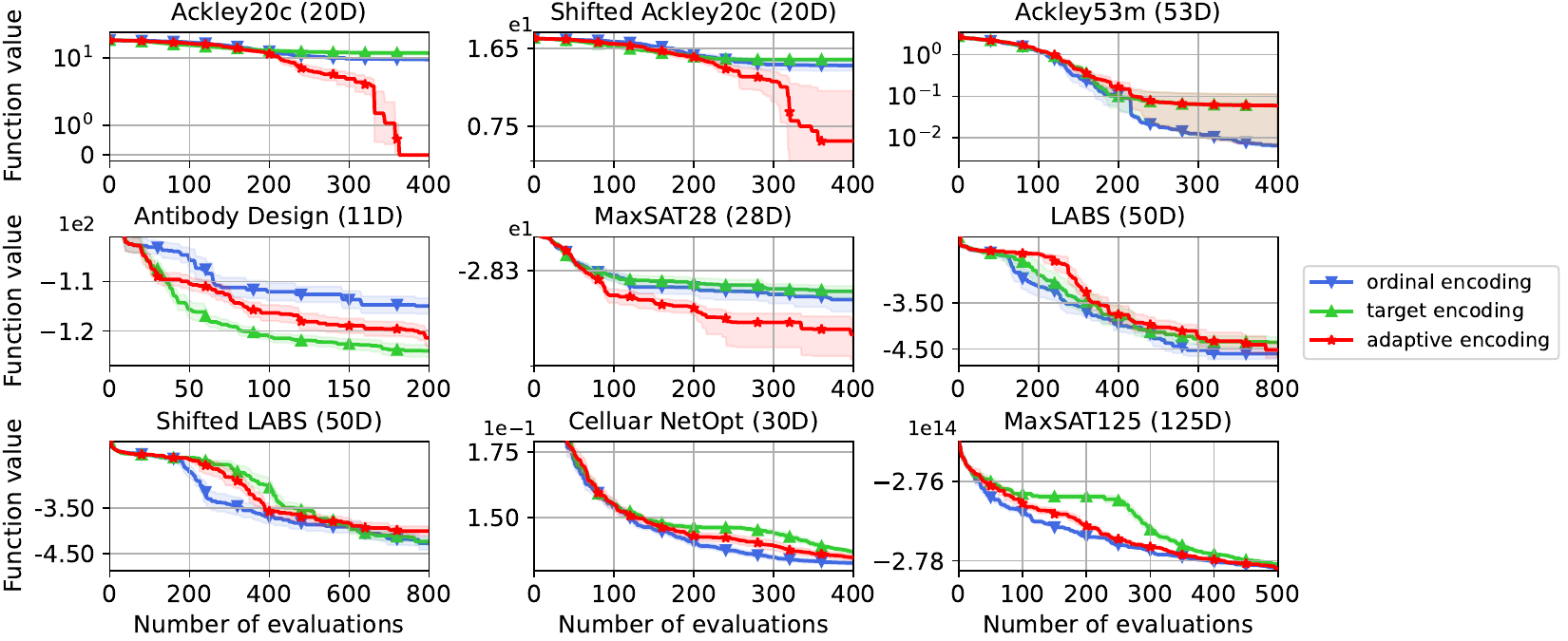}
\caption{The effectiveness of using the proposed \textit{adaptive} encoder selection mechanism against using only a single type of encoder. The performance of the adaptive encoder version is better or at least on par with the other versions.} \label{fig:exp3_test}
\end{figure*}

Fig.~\ref{fig:results} compares our proposed methods against all baselines. Firstly, we show that \textit{\cmabo, \cmacasmo and \cmabounce noticeably  outperform their respective BO optimizers}, demonstrating the effectiveness of our proposed \cmacma meta-algorithm. Specifically,
\cmabo consistently outperforms standard BO in all 9 benchmark problems by large margins. Notably, on Ackley20c, \cmabo surpasses all baselines and effectively finds the global optimum within the budget, highlighting the effect of \cmacma in identifying the global optimum.
Compared to CASMOPOLITAN, \cmacasmo shows superior performance on 7 out of the 9 benchmark problems. In these 7 problems, \cmacasmo consistently outperforms CASMOPOLITAN across all iterations. 
\cmabounce outperforms Bounce on 6 benchmark problems, including Shifted Ackley20c, Ackley53, LABS, Shifted LABS, MaxSAT28 and MaxSAT125. On MaxSAT28, LABS and Shifted LABS, \cmabounce maintains more optimal than Bounce across all iterations. On MaxSAT125D, \cmabounce initially performs similarly to Bounce but eventually finds more optimal solutions as the iterations progress.

\begin{figure*} [t]
 \centering
   \includegraphics[trim={0 0.5cm 0 0.5cm}, width=0.94\textwidth]{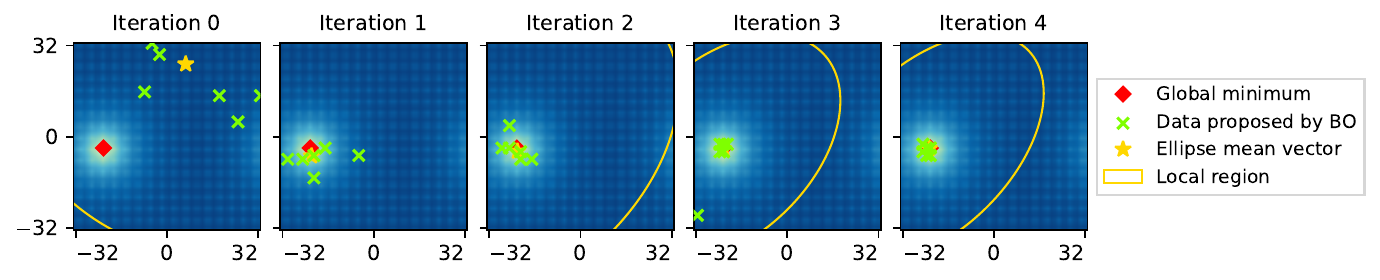}
\caption{The trajectories of the local regions from \cmabo on the 2D Shifted-Ackley categorical function. The hyper-ellipsoid local region starts at a random location, then evolves to better locations and finally directs the mean vector to the global optimum. The radii are prevented from shrinking too small to maintain a sufficient number of candidate points inside the local regions.} \label{fig:trajectory_sackley}
\end{figure*}

Finally, \textit{all three \cmacma methods outperform other baselines}, including TPE, SMAC, CMA-ES and Random Search on all 9 benchmark problems. These results showcase the effectiveness of the \cmacma meta-algorithm in enhancing the performance of state-of-the-art BO optimizers.

\subsection{Ablation Study and Analysis}

\paragraph{The Effectiveness of Adaptive Encoder Selection.} \label{sec-exp:adaptive_encoder_effectiveness}

We evaluate the effectiveness of our proposed categorical encoder selection in choosing the optimal encoder for the HESP strategy. We compare three variants of \cmabo, each using a different categorical encoder strategy: (1) the ordinal encoder, (2) the target encoder, and (3) our proposed adaptive optimal encoder selection. We perform the analysis on all 9 benchmark problems. 
In Fig.~\ref{fig:exp3_test}, we can see that the adaptive version has the best performance. The ordinal encoder shows good results on MaxSAT125 and LABS, yet falls behind substantially on Antibody Design. The target encoder is the most optimal on the Antibody Design problem, but degrades substantially on Ackley20c and MaxSAT125. The analysis indicates the robustness of our proposed adaptive encoder selection strategy for \cmacma.

\paragraph{The Trajectories of Local Regions under the \cmacma Meta-algorithm.}

We visualize the trajectories of the local regions generated by our \cmacma meta-algorithm on two 2D test functions: Ackley and Shifted Ackley. The search domain $[-32.768,\dots,32.768]$ is discretized into 51 evenly spaced points. Fig.~\ref{fig:trajectory_sackley} shows the trajectories for the Shifted Ackley function; additional plots for the Ackley function are provided in the 
Appendix~\ref{appendix:trajectory_plot}. 
Note that, we can only plot for \cmabo because the Hamming distance criteria in the local region of \cmacasmo and \cmabounce do not work in 2D problems (${L_h}_{\text{min}} = 1$ while ${L_h}_{\text{init}} = 2$). In Fig.~\ref{fig:trajectory_sackley}, we can see that at the beginning (Iteration 0), the proposed data points are scattered randomly to explore the region, as the prior information of BO is insufficient. However, when moving on to the next iterations, the local region evolves closer to the global optimum, helping BO to propose better solutions. Note that in combinatorial spaces, the search distribution is lower-bounded by a minimum standard deviation threshold in order to maintain a sufficiently large area for sampling~\cite{marty2023cmaes_benchmark}. Therefore, the radii of the local region are also retained large, while the mean vector gradually moves closer to the global minimum.


\section{Conclusion} \label{sec:conclusion}

In this paper, we propose \cmacma, the \textit{first} \textit{meta-algorithm} for high-dimensional BO in the \hybrid domains. \cmacma incorporates categorical encoders, which are optimally selected via MAB, to partition the mixed search space into  hyper-ellipsoid local regions for local modeling and optimization. We further derived three practical algorithms, namely \cmabo, \cmacasmo and \cmabounce, by incorporating \cmacma with the state-of-the-art BO optimizers for \hybrid variables: BO, CASMOPOLITAN, Bounce, respectively. Our extensive empirical results demonstrate the effectiveness of our proposed \cmacma meta-algorithm.



\begin{ack}
This work is partially supported by Australian Research Council (ARC) Discovery Project DP220103044. The first author (L.N.) would like to thank the School of Computing Technologies, RMIT University, Australia and the Google Cloud Research Credits Program for providing computing resources for this project. Additionally, this project was undertaken with the assistance of computing resources from RMIT AWS Cloud Supercomputing Hub (RACE).
\end{ack}



\bibliography{ecai2025}

\appendix
\section{Appendix} \label{appendix}
\subsection{General Process in Bayesian Optimization} \label{sec:appendix_bo}
BO addresses the optimization problem in an iterative fashion. At iteration $t$, BO approximates the objective function $f$ using a surrogate model based on the observed data collected so far $\mathcal{D}_t = \{\bm{z}_i, y_i\}|_{i=0}^t$, where $y_i = f(\bm{z}_i)+\varepsilon_i$ and $\varepsilon_i \sim \mathcal{N}(0, \sigma^2)$ represents noise. Subsequently, an acquisition function $\alpha: \mathcal{Z} \rightarrow \mathbb{R}$ is constructed to assign scores to data points in the domain $\mathcal{Z}$ based on their potential to identify a better solution. The next data point for observation is selected by maximizing the acquisition function $\bm{z}_{\text{next}}=\arg{\max\nolimits_{\bm{z}\in\mathcal{Z}}{\alpha(\bm{z})}}$ and is evaluated with the objective function $y_{\text{next}} = f(\bm{z}_{\text{next}}) + \varepsilon$, $\varepsilon \sim \mathcal{N}(0,\sigma^2)$. This newly observed data $\{\bm{z}_{\text{next}}, y_{\text{next}}\}$ is then aggregated to the current dataset. This process is conducted iteratively until the predefined budget, such as the maximum number of function evaluation, is depleted. Finally the best value found in the observed dataset is returned as the estimate of the global optimum $\bm{z}^*$.

\subsection{Surrogate Models and Acquisition Functions for Combinatorial and Mixed BO} \label{sec:appendix_bo_model_acq}

\paragraph{Surrogate Models.} In BO, the most popular surrogate model is Gaussian Process (GP) \cite{Rasmussen2006GP}, which depends on the kernel to capture the relationship between the variables. To capture the relationship between different ordinal/categorical variables and between ordinal/categorical and continuous variables, various GP kernels have been proposed, including Graph Diffusion Kernel \cite{ChangYong2019COMBO}, 
Additive Diffusion Kernel \cite{deshwal2021hybo}, 
Overlapped Kernel \cite{Bin2020CoCaBO}, 
Transformed Overlapped Kernel \cite{wan2021casmopolitan}, 
Tree Ensemble Kernel \cite{thebelt2022leafGP}, 
and String Kernel \cite{moss2020boss}. 
Beyond GP, there other surrogate models that can naturally represent the categorical data, such as Tree Parzen Estimator \cite{bergstra2011TPE} and Random Forest \cite{hutter2011smac}.

\paragraph{Acquisition Functions.} Acquisition functions for combinatorial and mixed BO methods are similar to those in continuous domains, such as Expected Improvement (EI) \cite{Mockus1978EI},
Upper Confidence Bound \cite{Srinivas2010UCB} 
and Thompson Sampling \cite{Thompson1933TS}.
In order to optimize these acquisition functions for combinatorial variables, many methods have been proposed, such as Simulated Annealing \cite{baptista2018bocs}, Local Search \cite{ChangYong2019COMBO,wan2021casmopolitan,papenmeier2023bounce}, Gradient-based Search \cite{wan2021casmopolitan,deshwal2023bodi,papenmeier2023bounce}. For mixed variables, Interleaved Search \cite{wan2021casmopolitan,papenmeier2023bounce,Bin2020CoCaBO} is often used to alter between optimizing ordinal/categorical and continuous acquisition functions.

\subsection{Categorical Encoders} \label{sec:appendix_categorical_encoder}
\paragraph{Ordinal Encoder.} 
Ordinal encoder is a 1-to-1 mapping $g: \mathcal{H} \rightarrow \mathbb{R}^{d_h}$ that assigns each categorical value with a unique numerical value.
Given a $c$-choice categorical variable $h \in \{u_1, u_2, \dots, u_c\}$, 
the ordinal encoded value of $u_i$ is $g_{\text{ordinal}}(u_i) = a_i$,
where $a_i \in \mathbb{R}$ and $a_i \neq a_j$ if $i\neq j$ for $i,j=1,\dots,c$. 
Ordinal encoder is a simple choice to transform ordinal/categorical values to numerical values, based on an existing or assumed relationships on the orders and rankings among ordinal/categorical values.

\paragraph{Target Encoder.}
Target encoder \cite{micci2001targetencoder} is a 1-to-1 mapping $g: \mathcal{H} \rightarrow \mathbb{R}^{d_h}$ that maps categorical variables to the statistics of the target variable. In practice, the most common statistics to be used for target encoder is the mean value. 
Denote a $c$-choice categorical variable $h \in \{u_1, u_2, \dots, u_c\}$. Let the observed dataset of $n$ samples with respect to the variable $h$ be denoted as $D=\{h_i, y_i\}_{i=1}^{n}$ where $y_i$ is the objective function value. The vector of target values associated with categorical value $u_i$ is defined as $\bm{y}_{u_i}=\{y_k | u_i = h_k, k =1,\dots,n\}$. Denote $n_i = \sum_{k=1}^{n}{\mathbb{I}(u_i, h_k)}$ as the frequency that $u_i$ exists in the dataset $D$, where $\mathbb{I}(.,.)$ is an indicator function.
The target encoded value of $u_i$ can be computed via its target mean value $\bar{y}_{u_i}=\sum_{k=1}^{n}{(y_i \mathbb{I}(u_i, h_k))}/ n_i$, such that, $g_{\text{target}}(u_i) = \frac{ n_i \bar{y}_{u_i} + m \bar{y} }{n_i + m}$,
where $\bar{y} = \sum_{i=1}^{n}{y_i}/ n $ is the overall mean of the dataset, and $m$ is a weight factor.

The main goal of target encoder is to map each the categorical value $u_i$ to its expected target value $\mathbb{E}[y_{u_i}|u_i]$. For large datasets, where categorical value $u_i$ appears frequently, i.e., $n_i$ is large, the encoded value $g_{\text{target}}(u_i)$ can be reasonably approximated by the target mean value $\bar{y}_{u_i}$. However, when $n_i$ is small, the target mean value can become unreliable due to small sample size \cite{micci2001targetencoder}. To address this, the overall mean value $\mathbb{E}[y] = \bar{y}$ is incorporated with a tuneable weighting factor $m$, which controls the extent to which the overall mean value is considered when $u_i$ does not have sufficient data. The target encoder formula is derived from the Empirical Bayes estimation theory \cite{robbins1992empiricalbayes}, which uses the empirical prior calculated from the observed data - the expected overall mean value $\mathbb{E}[y]$ - to adjust the empirical posterior computed from the observed data for each category $u_i$ - the expected target value $\mathbb{E}[y_{u_i}|u_i]$.

\subsection{CASMOPOLITAN} \label{sec:appendix_casmo}
CASMOPOLITAN \cite{wan2021casmopolitan} is a state-of-the-art high-dimensional BO method for categorical and mixed spaces. To tackle the high dimensions, it employs a local search strategy to define small local regions, called Trust Region (TR) \cite{Eriksson2019TuRBO} and performs BO within. CASMOPOLITAN deals with categorical and continuous variables separately by maintaining two respective TRs. The continuous TR is a hyper-rectangle centered at the best solution found so far, with its side lengths determined by the GP length-scales multiplied with a length ratio factor $L_x$. The categorical TR is constructed as a region centered at the best solution found so far, and contain the data points within a Hamming distance of $L_h$ to the TR center. During optimization, CASMOPOLITAN adaptively expands or shrinks the TRs (increases or decreases both $L_x$ and $L_h$) depending on the success or failure of the algorithm. 

When modelling the GP, CASMOPOLITAN proposes to use the Transformed Overlapped Kernel, which can capture the relationship between categorical variables, and also relationship between categorical and continuous variables. Because this kernel uses Hamming distance, CASMOPOLITAN simply uses ordinal encoding to transform categorical data into numerical data for GP modelling.

\subsection{Bounce} \label{sec:appendix_bounce}
Bounce \cite{papenmeier2023bounce} is a state-of-the-art high-dimensional BO method for combinatorial and mixed spaces. To tackle the high dimensionality,
Bounce employs a subspace embedding method \cite{Nayebi2019HighdimBOEmb, Letham2020Alebo,papenmeier2022baxus}, which leverages the random linear embedding to map the $d$-dimensional input vectors $\mathbf{z} \in \mathcal{Z}$ to $d_\mathcal{A}$-dimensional vectors in a low-dimensional target space $\mathcal{A}$, meaning $d_\mathcal{A}<d$. This random embedding allows for optimization in a low-dimensional subspace $\mathcal{A}$. During optimization, Bounce adaptively increases the target space dimensionality $d_\mathcal{A}$ to allow greater flexibility compared to the fixed choice of $d_\mathcal{A}$ as in previous methods \cite{Nayebi2019HighdimBOEmb,Letham2020Alebo}. In ordinal, categorical and mixed settings, Bounce only maps variables of the same type together using the same embedding rules. Bounce also employs the local search strategy and maintains two different TRs (with factors $L_x$ and $L_h$), similar to CASMOPOLITAN. However, the TR management rule of Bounce is different compared to other TR-based methods \cite{Eriksson2019TuRBO,papenmeier2022baxus,wan2021casmopolitan}. Bounce instantly changes the TR sizes (both $L_h$ and $L_x$) after each iteration, while other methods require the algorithms to maintain consecutive successes (or failures) to change the TR sizes. 

When modelling the GP, Bounce uses one-hot encoder to transform categorical variables to approximate the GP. Bounce leverages the Overlapped Kernel \cite{Bin2020CoCaBO} by combining two separate Matern-5/2 kernels to capture the relationship between categorical and continuous variables.

\subsection{Experiment Setup} \label{appendix:experiment_setup}
For GP modelling and acquisition function choice, in all derived methods of \cmacma, we use similar surrogate model and acquisition functions as in the respective BO optimizers.

For the HESP mechanism, we use the code implementation that is made available\footnote{\url{https://github.com/CMA-ES/pycma}} to compute and update the search distribution. We follow \cite{ngo2024cmabo,Hansen2016CMAESTutorial} to set the population size $\lambda$. We further set the lower bound for the standard deviations of ordinal and categorical dimensions, as suggested in \cite{marty2023cmaes_benchmark}. When initializing the search distribution from the observed dataset ${\mathcal{D}_0}_e$, we set the mean vector $m^{(0)}$ as the minimizer of the dataset, the covariance matrix $\mathbf{\Sigma}^{(0)}$ as the identity matrix, and the step-size $\sigma=0.3[u-l]$, where $[l,u]^d$ is the boundary region. 

For the categorical encoders, we use the package \texttt{category-encoders}\footnote{\url{https://pypi.org/project/category-encoders/}} of \texttt{scikit-learn} and keep the factor $m$ as default. 

For the adaptive encoder selection, we follow \cite{auer2002exp3} to set $\eta=\min\{1, \sqrt{K\ln{K} / ((1-e)N)}\}$, where $N$ is the maximum number of iteration budget.

We implement \cmacma in Python (version 3.10). We provide in the code supplementary a \texttt{.yml} file to install the required packages for running our proposed methods.

\subsection{Baselines} \label{appendix:baseline_details}
To evaluate the baseline methods, we use the implementation and hyper-parameter settings provided in their public source code and their respective papers. All the methods are initialized with 20 initial data points and are run for 10 repeats with
different random seeds. All experimental results are averaged over these 10 independent runs. We then report the best minimum value found. 

\paragraph{Standard BO.} This is the standard BO method with the TS acquisition function. We apply an ordinal encoder to transform ordinal and categorical variables into continuous variables, which are then modelled via continuous kernels. Specifically, we use the Matern 5/2 ARD kernel to model the transformed continuous variables.


\paragraph{CASMOPOLITAN \cite{wan2021casmopolitan}.} We set all the hyper-parameters of CASMOPOLITAN as suggested in their paper. 
This includes the upper and lower bound for both TR side length ${L_x}_{\max}=1.6$, ${L_x}_{\min}=2^{-7}$, ${L_h}_{\max}=d$, ${L_h}_{\min}=1$ and the TR adaptation threshold $\tau_{\text{succ}}=3$, $\tau_{\text{fail}}=40$ where $d$ is the dimension of the problem. For GP modelling, we use Transformed Overlapped kernel. For the input encoder, we use an ordinal encoder. For the acquisition function, we use TS \cite{Thompson1933TS}. We use their implementation that is made available at \url{https://github.com/xingchenwan/Casmopolitan}.

\paragraph{Bounce \cite{papenmeier2023bounce}.} We set all the hyper-parameters of Bounce as suggested in their paper. 
This includes the upper and lower bound for both TR side length ${L_x}_{\max}=1.6$, ${L_x}_{\min}=2^{-7}$, ${L_h}_{\max}=d_\mathcal{A}$, ${L_h}_{\min}=1$ and $m_D$ is set to half of the maximum budget.
For GP modelling, we use the weighted combination of Matern 5/2 kernels via addition and multiplication. For input encoder, we use one-hot encoder. For acquisition function, we use EI \cite{Mockus1978EI} which is optimized via Local Search (or Interleaved Search for mixed cases).
We use their implementation that is made available at \url{https://github.com/LeoIV/bounce}.

\paragraph{CMAES \cite{Hansen2001CMAES}.} We use the default settings as suggested in the paper. This includes the population size $\lambda=4+\lfloor3+\ln d\rfloor$ where $d$ is the problem dimension, the random initial mean vector $\bm{m}^{(0)}$ selected from the minimum of the 20 random initial points, the identity initial covariance matrix $\bm{C}^{(0)} = \bm{I}_{d}$ and the initial step-size $\sigma^{(0)}=0.3(u-l)$ where the domain is scaled to uniform bound of $[l,u]^d$. We also activate the restart mechanism of CMA-ES so that the algorithm can restart when it converges to a local minimum. For the input encoder, we use an ordinal encoder. To handle ordinal and categorical variables, we follow \citeauthor{marty2023cmaes_benchmark} and set the lower bound of the standard deviations of the ordinal and categorical dimensions $\sigma_{\text{LB}} = 0.1$
We use their implementation that is made available at \url{https://github.com/CMA-ES/pycma}.

\paragraph{TPE \cite{bergstra2011TPE}.} We use the implementation and default settings from \texttt{hyperopt} Python package. The code is available at \url{http://hyperopt.github.io/hyperopt/}.

\paragraph{SMAC \cite{hutter2011smac}.} We use the implementation and default settings from \url{https://github.com/automl/SMAC3}.

\subsection{Experiments Details} \label{appendix:experiment_details}


We additionally provide more information on the test problems below. See a summary in Table \ref{table:test_problems}.
\paragraph{Synthetic Benchmark Functions.}
The problem Ackley20c is the common Ackley function\footnote{\url{https://www.sfu.ca/~ssurjano/ackley.html}} with 20 dimensions, in which each dimension is evenly discretized into 11 values within the interval $[-32.768,32.768]$. This implementation is from \cite{dreczkowski2024mcbo}. The problem Ackley53m is also an Ackley function with 53 dimensions, which consists of 50 binary variables $[0, 1]^{50}$ mapped to the boundary $[0, 1]^{50}$, and 3 continuous variables $[0, 1]^3$. This implementaion has been used in \cite{wan2021casmopolitan,deshwal2023bodi,papenmeier2023bounce}.

\paragraph{Real-world Benchmark Problems.}
The problem Antibody Design finds a string sequence representing the antibody to optimize a binding energy between the antibody and an antigen. The string sequence is represented as a 11D vector of categorical variables, each has 20 different choices. We use the implementation in \cite{dreczkowski2024mcbo}.
The problem LABS (Low-Autocorrelation Binary Sequences) finds a binary sequence to maximize a merit factor. The sequence is represented as a 50D vector of binary variables. We use the implementation in \cite{deshwal2023bodi,papenmeier2023bounce}.
The problems MaxSAT28 and MaxSAT125 (maximum satisfiability problem) \cite{hansen1990maxsat} find a 28-dimensional and 125-dimensional binary vector to maximize the combined weights of satisfied clauses. We use the implementation in \cite{ChangYong2019COMBO,wan2021casmopolitan,deshwal2023bodi,papenmeier2023bounce}.
The problem Cellular Network has 30 dimensions mixed between 15 continuous values to represent the transmission power and 15 ordinal variables (each with 6 values) to represent the tilting angle of the antennas. The goal is to maximize the coverage quality of the antenna system. We use the implementation in \cite{daulton2022bopr}.

For the shifted functions, we follow \citeauthor{papenmeier2023bounce} \cite{papenmeier2023bounce} to permute the categories of the categorical variables by a uniformly random vector $\bm{\delta} = [\delta_1,\dots,\delta_{d_\mathcal{H}}]$ and $\delta_i \sim \mathcal{U}(1,\dots, c_i) \in \mathbb{N}$ where $c_i$ is the number of choices for categorical variables  $h_i$. Therefore, the shifted functions are defined as $f_\text{shifted}(\bm{h}) = f_\text{original}((\bm{h} + \bm{\delta})\mathbin{\%} \bm{c})$, where $\bm{c} = [c_1,\dots,c_{d_\mathcal{H}}]$ and $\mathbin{\%}$ is the modulo operator. The motivation for these shifted functions are to remove special structure of the global optimum \cite{papenmeier2023bounce,ngo2024cmabo}.


\subsection{Additional Trajectory Plot} \label{appendix:trajectory_plot}
\begin{figure*} [t]
  \centering
    \includegraphics[width=\textwidth]{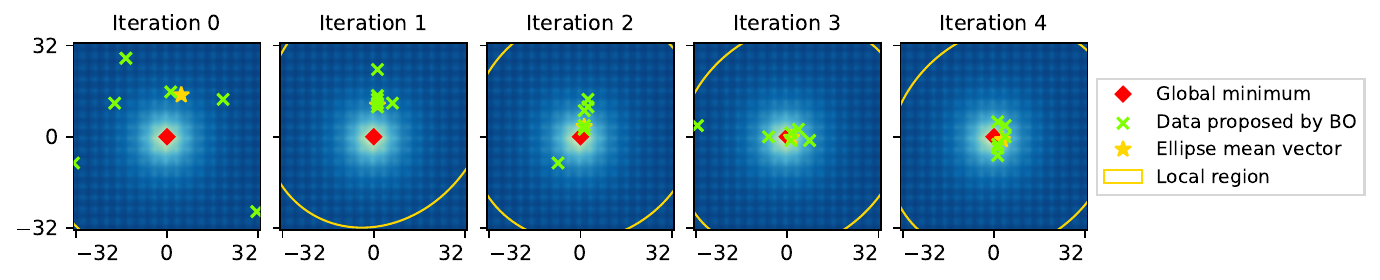}
\caption{The trajectory of the local regions from \cmabo on 2D Ackley categorical function.} \label{fig:trajectory_ackley}
\end{figure*}

We present the trajectory plot for Ackley-2D function in Fig. \ref{fig:trajectory_ackley}.

\subsection{Computing Infrastructure}
We run experiments on a server with a CPU of type AMD EPYC 7R32 (clock speed 3.3GHz). Each experiments are allocated 8 CPUs and 32GB Memory. 
The server is installed with Operating System Ubuntu 20.04.3 LTS. We use Miniconda (version 23.3.1) to install Python packages.

\subsection{Running Time} We report the running time per iteration (in seconds) of our proposed methods and baselines in Table \ref{table:running_time}. We also report in Table \ref{table:running_time_ablation} the running time per iteration (in seconds) of MOCA-HESP-BO when using adaptive encoders and fixed encoders (ordinal, target). The results in Table \ref{table:running_time_ablation} confirm that adaptive selection incurs minimal computational time.

\begin{table*} [hb]
\caption{Average running time per iteration (in seconds).}
\begin{center}
\begin{tabular}{lcccccccc}
\hline
 & TPE & SMAC & BO & \makecell[c]{MOCA-\\HESP-\\BO} & \makecell[c]{CASMO-\\POLITAN} & \makecell[c]{MOCA-\\HESP-\\Casmo} & Bounce & \makecell[c]{MOCA-\\HESP-\\Bounce} \\
\hline
Ackley20c & 0.03 & 0.27 & 0.64 & 0.80 & 4.82 & 6.15 & 1.23 & 2.73 \\
Antibody Design & 0.01 & 0.23 & 0.64 & 3.23 & 6.33 & 10.82 & 1.75 & 3.68 \\
LABS & 0.06 & 0.41 & 2.90 & 2.46 & 8.90 & 9.48 & 2.19 & 3.46 \\
MaxSAT28 & 0.03 & 0.08 & 0.63 & 0.86 & 5.49 & 4.06 & 0.72 & 1.28 \\
MaxSAT125 & 0.41 & 1.00 & 2.73 & 3.83 & 11.78 & 19.65 & 3.71 & 4.44 \\
Shifted Ackley20c & 0.03 & 0.27 & 0.70 & 0.79 & 5.04 & 6.04 & 1.25 & 2.70 \\
Shifted LABS & 0.06 & 0.41 & 2.80 & 2.35 & 9.32 & 9.53 & 2.50 & 3.44 \\
Ackley53m & 0.05 & 0.31 & 1.74 & 2.23 & 18.07 & 18.17 & 3.75 & 15.87 \\
Cellular Network & 0.02 & 0.30 & 1.20 & 1.10 & 7.58 & 8.40 & 8.45 & 17.98 \\
\hline
\end{tabular} \label{table:running_time}
\end{center}
\end{table*}

\begin{table*} [hb]
\caption{Average running time per iteration (in seconds) of MOCA-HESP-BO when using adaptive encoders and fixed encoders (ordinal, target). The results confirm that adaptive selection incurs minimal computational time.}
\begin{center}
\begin{tabular}{lcccccccc}
\hline
 & Adaptive & Ordinal & Target \\
\hline
Ackley20c & 0.80 & 0.62 & 0.66 \\
Antibody Design & 3.23 & 1.99 & 1.81 \\
LABS & 2.46 & 2.07 & 2.20 \\
MaxSAT125 & 3.83 & 3.92 & 3.49 \\
Shifted Ackley20c & 0.79 & 0.64 & 0.65 \\
Shifted LABS & 2.35 & 1.99 & 1.80 \\
Ackley53m & 2.23 & 1.84 & 1.92 \\
Cellular Network & 1.10 & 1.13 & 0.97 \\
\hline
\end{tabular} \label{table:running_time_ablation}
\end{center}
\end{table*}

\end{document}